\documentclass[dvipsnames]{article}
\usepackage{mlconf,times}

\usepackage{hyperref}
\usepackage{url}
\usepackage{subcaption}

\usepackage{cite}
\usepackage{amsmath,amssymb,amsfonts}
\usepackage{algorithm}
\usepackage{algpseudocode}

\usepackage{graphicx}
\usepackage{textcomp}
\usepackage{xcolor}

\usepackage{listings}
\usepackage{xspace}
\usepackage{soul}
\usepackage{annotate-equations}
\usepackage{ctable}
\usepackage{enumitem}
\usepackage{todonotes}

\renewcommand{\eqnannotationfont}{\bfseries\small}

\newcommand{\shcolor}{PineGreen}
\newcommand{\jrcolor}{NavyBlue}
\newcommand{\shalg}[1]{\colorbox{\shcolor!17}{$\displaystyle #1$}}
\newcommand{\shtxt}[1]{\colorbox{\shcolor!17}{#1}}
\newcommand{\jralg}[1]{\colorbox{\jrcolor!17}{$\displaystyle #1$}}
\newcommand{\jrtxt}[1]{\colorbox{\jrcolor!17}{#1}}

\def\BibTeX{{\rm B\kern-.05em{\sc i\kern-.025em b}\kern-.08em
    T\kern-.1667em\lower.7ex\hbox{E}\kern-.125emX}}

\definecolor{dkgreen}{RGB}{0,64,0}
\definecolor{ltgray}{RGB}{245,245,245}
\definecolor{mauve}{RGB}{139,0,139}

\newcommand{\tweakedsim}{\raise.17ex\hbox{$\scriptstyle\mathtt{\sim}$}}

\newcommand\jorge{Jorge\xspace}

\title{\jorge: Approximate Preconditioning for GPU-efficient Second-order
Optimization}

\author{Siddharth Singh, Zachary Sating, Abhinav Bhatele\\
Department of Computer Science, University of Maryland\\
ssingh37@umd.edu, zsating@umd.edu, bhatele@cs.umd.edu
}

\mlconffinalcopy
\begin{document}

\maketitle

\begin{abstract}
Despite their better convergence properties compared to first-order optimizers,
second-order optimizers for deep learning have been less popular due to their
significant computational costs. The primary efficiency bottleneck in such
optimizers is matrix inverse calculations in the preconditioning step, which
are expensive to compute on GPUs. In this paper, we introduce \jorge, a
second-order optimizer that promises the best of both worlds -- rapid
convergence benefits of second-order methods, and high computational efficiency
typical of first-order methods. We address the primary computational bottleneck
of computing matrix inverses by completely eliminating them using an
approximation of the preconditioner computation. This makes \jorge extremely
efficient on GPUs in terms of wall-clock time. Further, we describe an approach
to determine \jorge's hyperparameters directly from a well-tuned SGD baseline,
thereby significantly minimizing tuning efforts. Our empirical evaluations
demonstrate the distinct advantages of using \jorge, outperforming
state-of-the-art optimizers such as SGD, AdamW, and Shampoo across multiple
deep learning models, both in terms of sample efficiency and wall-clock time.

\end{abstract}

\section{Introduction}
\label{sec:intro}
Stochastic optimization methods such as stochastic gradient descent
(SGD)~\citep{sgd} and {\em Adam}~\citep{KingmaAdam2014} are the de-facto
standard for optimizing the objective function in the training of deep neural
networks. These first-order optimization methods are relatively inexpensive in
terms of their compute and memory requirements, and hence extremely popular.
Second-order optimization methods typically have better convergence properties
(fewer epochs to reach target validation metrics) than those of first-order
methods. However, they are considerably slower in terms of per-iteration
(per-batch) wall-clock times for training than first-order methods. This is
because they often use a preconditioner, which multiplies the gradient by a
matrix before taking a step. Computing these preconditioners requires
performing matrix inversions, which are highly inefficient on GPU platforms due
to the iterative nature of matrix inverse algorithms and their irregular memory
access patterns.

If one could develop a second-order optimizer that has better convergence than
first-order methods and is on par with them in terms of wall-clock time per
iteration, we could achieve the best of both worlds. In this paper, we present
\jorge\footnote{\jorge is named after Dr.~Jorge Nocedal, an applied
mathematician \& expert in nonlinear optimization.}, a new second-order
optimizer that uses an approximation for preconditioning by avoiding the
calculation of the inverse of matrices in all steps. It has similar convergence
properties to other second-order optimization methods but its wall-clock time
per iteration is similar to that of inexpensive first-order methods. This is a
win-win situation, which leads to much faster total training times for several
different deep learning models when compared to other state-of-the-art
optimizers.

A new optimization method is most useful and promising if users do not have to
spend significant time in tuning its hyperparameters. We demonstrate the
process of deriving reasonable hyperparameters for \jorge from a well-tuned SGD
baseline with minimal effort. Interestingly, these derived hyperparameters
match the generalization of SGD and even improve it in many cases! Note that we
use SGD over other adaptive optimizers such as Adam because prior research has
shown that SGD often outperforms adaptive methods in terms of
generalization~\citep{adapative-optimizer-bad-sgd-good}. In our experiments
across different network architectures, we demonstrate that \jorge performs
better than two widely adopted first-order optimizers, SGD and AdamW, both in
terms of sample efficiency and overall wall-clock times for convergence.
Additionally, we demonstrate comparable sample efficiency to
Shampoo~\citep{shampoo-icml}, a state-of-the-art second-order optimizer, while
achieving faster convergence times.

\subsection{Contributions}

This paper makes the following important contributions:
\begin{itemize}[noitemsep, nolistsep, leftmargin=10pt]
\item A new second-order optimizer that avoids matrix inverse calculations when
computing the preconditioner, making it extremely efficient on GPUs.  This
results in per-iteration wall-clock times within 5-10\% of those of first-order
optimizers such as SGD and AdamW, while matching the sample efficiency of
Shampoo, a second-order optimizer. For training ResNet-50 on ImageNet, we
demonstrate improvements of nearly 25\% in the total training wall-clock time
over SGD. 
\item We show that reasonable hyperparameter configurations for \jorge can be
easily bootstrapped from those of a well-tuned SGD baseline without extensive
hyperparameter tuning that would require full training runs. These settings
result in either similar and in many cases, even better generalization than
that of SGD!
\item Most second-order optimizers need to exploit complex parallelism
requiring multiple GPUs to get their total training times to be faster than
those of first-order optimizers. Since \jorge is highly efficient, it can be
run locally on each GPU and still outperform highly optimized parallel
implementations of second-order optimizers.
\end{itemize}

\subsection{Related work}
\label{sec:related}

There have been several research efforts to develop computationally tractable
second-order optimizers for deep learning. \citet{hessian-free-optimization}
proposes Hessian-free optimization, which exploits conjugate gradient (CG) to
directly compute Hessian-vector products without explicitly computing the
Hessian. Since CG requires multiple iterations, there has been subsequent work
on reducing this cost~\citep{erdogdu2015convergence}. Several optimizers based
on the L-BFGS method have also been proposed that approximate Hessian-vector
products from the history of past gradients, again without explicitly computing
the Hessian~\citep{berahas2016multibatch, bollapragada2018progressive,
wang2017stochastic}. 

Most state-of-the-art second-order optimizers rely on block-diagonal
approximations of the Hessian to reduce the computational and memory
requirements. The ``blocks'' typically correspond to substructures in the
neural network, like a layer or a parameter tensor. Some recent methods in this
category include Shampoo~\citep{shampoo-icml}, K-FAC~\citep{kfac:2015,
grosse2016kfacconvolution}, K-BFGS~\citep{goldfarb2020practical} and the GGT
method~\citep{agarwal-ggt-2019}. However, these methods need to compute the
inverse of their approximate Hessian matrices, which can be expensive to
compute even with the block-diagonal approximations. As we show later in
Section~\ref{sec:exp-results}, \jorge outperforms one such optimizer, Shampoo,
by nearly 37\% in terms of the total wall-clock time for training ResNet-50 on
ImageNet. Closely related to \jorge is a line of work that exploits the
Sherman-Morrison based Matrix identity to approximate the update steps in K-FAC
without computing any matrix inverses~\citep{mozaffari2023mkor, zhang2023eva,
tang2021skfac}.

To mitigate the large computational costs of matrix inverses, researchers have
also proposed parallel implementations of second-order optimizers, which aim to
distribute the work of the optimizer across multiple GPUs.  Several efforts
focus on developing efficient parallel implementations of the K-FAC
optimizer~\citep{pauloski-2020-kfac, pauloski-kaisa-2021, osawa2019largescale,
osawa2020scalable, ueno2020rich, shi2021accelerating}. On the other hand,
\citet{shi2023distributed} and \citet{shampoo-scalable} aim to accelerate the
Shampoo~\citep{shampoo-icml} optimizer via parallelism.
\citet{shampoo-scalable} present a heterogeneous solution that offloads the
computation of the inverses to the CPU. Even though we implement \jorge without
any multi-GPU parallelism, we demonstrate that its performance is better than
one of the state-of-the art parallel optimizers -- Distributed
Shampoo~\citep{shi2023distributed}.

\section{Background}
\label{sec:bg}
Second-order optimizers make use of both the gradients and curvature (second
derivatives) of the loss function. By considering the curvature, second-order
methods can approximate the loss function more accurately than first-order
optimizers, and thus reduce the number of iterations required for convergence.
Most second-order optimizers approximate the Newton step shown in
Equation~\ref{eqn:sorder}.

\vspace{0.1in}
\begin{equation}
    \eqnmarkbox[Mulberry]{T1}{\theta_{t}} = \eqnmarkbox[Mulberry]{T2}{\theta_{t-1}} - \eqnmarkbox[OliveGreen]{H1}{H^{-1}_{t}} \eqnmarkbox[NavyBlue]{G1}{G_{t}}
\label{eqn:sorder}
\end{equation}
\annotatetwo[yshift=1em]{above}{T1}{T2}{parameters at timestep t and t-1}
\annotate[yshift=-1em]{below,left}{H1}{Hessian at timestep t}
\annotate[yshift=-1em]{below,right}{G1}{gradients at timestep t}
\vspace{0.15in}

This equation can be derived by minimizing a second-order Taylor's
approximation of the loss function at $\theta_{t}$.  This step of multiplying
the gradients with $H^{-1}_{t}$ is called preconditioning, and $H^{-1}_{t}$ is
often referred to as a preconditioner.

Instead of using the actual Hessian, optimizers typically use positive
semi-definite approximations of the Hessian~\citep{schraudolphGGN,ngd-og} to
account for the non-convexity of the training
objective~\citep{krylov-subspace-descent, botev-practical-17, tonga, kfac:2015,
Desjardins:nips2015}. Our proposed optimizer, \jorge, belongs to a class of
methods called ``adaptive optimizers'', which use the inverse of the gradient
covariance matrix (or the empirical Fisher matrix) to precondition gradients.
Examples of adaptive second-order optimizers include the full matrix version of
Adagrad~\citep{duchi:jmlr2011} and Shampoo~\citep{shampoo-icml}. Note that
several first-order adaptive optimizers have also been proposed in literature,
which only use the diagonal elements of the covariance matrix. Popular examples
include Adam~\citep{KingmaAdam2014} and RMSProp. \citet{jastrzebski2018factors,
sagun2018empirical, zhu2019anisotropic} provide justification for the usage of
the gradient covariance matrix as an approximation of the Hessian.

\section{Approximate Preconditioning in \jorge}
\label{sec:formulation}
As described in Section~\ref{sec:related}, the primary efficiency bottleneck in
state-of-the-art second-order optimizers such as K-FAC~\citep{kfac:2015} and
Shampoo~\citep{shampoo-icml} is the matrix inverse computations performed to
calculate the preconditioners. To overcome this limitation, we introduce Jorge,
an efficient, adaptive, second-order optimizer tailored for GPU execution.
Jorge's formulation eliminates computing explicit matrix inversions, and is
solely comprised of matrix multiplications and additions, which are highly
optimized on GPUs. This results in Jorge's wall-clock time per iteration to be
on par with those of first-order optimizers, while also having faster
convergence properties typical of a second-order optimizer.

We propose Jorge as an enhancement of Shampoo~\citep{shampoo-icml}, another
adaptive second-order optimizer. We first describe Shampoo's optimizer
algorithm at a high level before describing \jorge's optimizer algorithm. Note
that, throughout this section, we discuss Shampoo and by extension Jorge,
within the context of a single layer. Application to multiple layers simply
involves repeating the same steps for their parameters.

Following~\citet{shampoo-icml}, let us assume that the parameters, $\theta$, of
a single layer are organized in a two-dimensional (2D) $\mathit{m \times n}$
matrix (N-dimensional parameter tensors, like those found in convolution layers
are typically collapsed into 2D matrices, in practice). Shampoo maintains the
second-order curvature information of the loss in two matrices -- $L_{t}$ (size
$m \times m$) and $R_{t}$ (size $n \times n$), which are called the left and
right preconditioners, respectively. It iteratively updates the preconditioners
from the current gradient information as shown in the equation below (for the
left preconditioner):

\vspace{0.2in}
\begin{equation}
    \eqnmarkbox[PineGreen]{L1}{L_{t}} = \eqnmarkbox[Mahogany]{b1}{\beta_{2}} \eqnmarkbox[PineGreen]{L2}{L_{t-1}} + (1-\beta_{2}) \eqnmarkbox[NavyBlue]{G1}{G_{t}} G^{T}_{t}
\label{eqn:shampoo-bottleneck}
\end{equation}
\annotatetwo[yshift=1em]{above}{L1}{L2}{left preconditioner at timestep t and t-1}
\annotate[yshift=-1em]{below,right}{G1}{gradients at timestep t}
\annotate[yshift=-1em]{below,left}{b1}{smoothing parameter}
\vspace{0.1in}

Algorithm~\ref{algorithm:shampoo} shows how the preconditioners are used in
Shampoo. Additional terms used in the algorithm are defined as follows.
$\beta_{1}$ and $\beta_{2}$ are smoothing parameters for the exponential moving
average (EMA) of the momentum and preconditioners. $\tilde{G_{t}}$ is the
preconditioned gradients at timestep $t$. $m_{t}$ is the EMA of the
preconditioned gradients, and $\eta_{t}$ is the learning rate at timestep $t$.
Lines 5--8 of Algorithm~\ref{algorithm:shampoo} show how the Shampoo optimizer
iteratively updates the left and right preconditioners from the current
gradients' information. Line 11 illustrates the preconditioning step, wherein
the gradients are multiplied by $L^{\frac{-1}{4}}_{t}$ and
$R^{\frac{-1}{4}}_{t}$ on the left and right, respectively. The preconditioning
step produces the preconditioned gradients, $\tilde{G}_{t}$, which minimize the
loss faster than the raw gradients.  Finally, we update the momentum estimate
of the preconditioned gradients (line 14), and then use the momentum to update
the weights (line 15).  The matrix inverse computation in the preconditioning
step (line 11) is the primary efficiency bottleneck in Shampoo, and is exactly
what we want to optimize in Jorge.

\begin{minipage}{0.35\textwidth}
\begin{algorithm}[H]
    \centering
    \caption{\shtxt{Shampoo}}\label{algorithm:shampoo}
    \begin{algorithmic}[1]
    {\footnotesize
        \State \textbf{Initialize} $\theta_{0}$, \shalg{L_{0}=\epsilon I_{m}}
        \State \hspace{0.72in} \shalg{R_{0}=\epsilon I_{n}}
        \For {t=1 ,..., T}
        \State \textbf{Update Preconditioners:}
        \State \shalg{L_{t} = \beta_{2} L_{t-1}}
        \State \hspace{0.32in} \shalg{+ (1-\beta_{2}) G_{t} G^{T}_{t}}
        \State \shalg{R_{t} = \beta_{2} R_{t-1}}
        \State \hspace{0.32in} \shalg{+ (1-\beta_{2}) G^{T}_{t} G_{t}}
        \State
        \vspace{0.39in}
        \State \textbf{Precondition Gradients:}
        \State \shalg{\tilde{G_{t}} = L^{\frac{-1}{4}}_{t} G_{t} R^{\frac{-1}{4}}_{t}}
        \State
        \State \textbf{Update Weights:}
        \State $m_{t} = \beta_{1} m_{t-1} + (1-\beta_{1}) \tilde{G_{t}}$ 
        \State $\theta_{t} = \theta_{t-1} - \eta_{t} m_{t}$
        \EndFor
    }
    \end{algorithmic}
\end{algorithm}
\end{minipage}
\hfill
\begin{minipage}{0.64\textwidth}
\begin{algorithm}[H]
    \centering
    \caption{\jrtxt{\jorge} compared to Shampoo}\label{algorithm:jorge}
    \begin{algorithmic}[1]
    {\footnotesize
        \State \text{\textbf{Initialize} $\theta_{0}$, \jralg{\hat{L}_{0}=\epsilon^{-\frac{1}{4}} I_{m}}, \jralg{\hat{R}_{0}=\epsilon^{-\frac{1}{4}} I_{n}}}
        \State
        \vspace{0.02in}
        \For {t=1 ,..., T}
        \State \textbf{Update Preconditioners:}
        \State \jralg{X_{L} = \hat{L}_{t-1}^{4} G_{t} G^{T}_{t}} 
        \State {\footnotesize \jralg{\hat{L}_{t} = \beta^{\frac{-1}{4}}_{2} \hat{L}_{t-1} \left( I_{m} - \frac{ (1-\beta_{2})}{4 \beta_{2}} X_{L} + \frac{5 (1-\beta_{2})^{2}}{32 \beta^{2}_{2}} X^{2}_{L} \right)}}
        \State \jralg{X_{R} = \hat{R}_{t-1}^{4}G^{T}_{t} G_{t}} 
        \State {\footnotesize \jralg{\hat{R}_{t} = (\beta^{\prime}_{2})^{\frac{-1}{4}}\hat{R}_{t-1} \left( I_{n} - \frac{ (1-\beta^{\prime}_{2})}{4 \beta^{\prime}_{2}} X_{R} + \frac{5 (1-\beta^{\prime}_{2})^{2}}{32 (\beta^{\prime}_{2})^{2}} X^{2}_{R} \right)}}
        \State
        \State \textbf{Precondition Gradients:}
        \State \jralg{\tilde{G_{t}} = \hat{L}_{t} G_{t} \hat{R}_{t}}
        \vspace{0.05in}
        \State
        \State \textbf{Update Weights:}
        \State $m_{t} = \beta_{1} m_{t-1} + (1-\beta_{1}) \tilde{G_{t}}$ 
        \State $\theta_{t} = \theta_{t-1} - \eta_{t} m_{t}$
        \EndFor
    }
    \end{algorithmic}
\end{algorithm}
\end{minipage}
\vspace{0.1in}

In Algorithm~\ref{algorithm:jorge}, we show the functioning of \jorge
side-by-side with Shampoo for the same 2D $\mathit{m \times n}$  parameter
matrix of a single layer. The core idea behind Jorge is to approximate the
computation of $L^{\frac{-1}{4}}_{t}$ and $R^{\frac{-1}{4}}_{t}$ in Shampoo
(line 11 of Algorithm~\ref{algorithm:shampoo}) in a GPU-efficient manner. In
order to do this, we modify the computation in both lines 5--8 and line 11 of
Algorithm~\ref{algorithm:shampoo}.  Just like Shampoo, Jorge also maintains two
preconditioners, which we refer to as $\hat{L}_{t}$ and $\hat{R}_{t}$ in
Algorithm~\ref{algorithm:jorge}. However, Jorge's preconditioners are an
approximation of the inverse fourth root of Shampoo's preconditioners at every
iteration, i.e. $\hat{L}_{t} \approx {L}^{\frac{-1}{4}}_{t}$ and $\hat{R}_{t}
\approx {R}^{\frac{-1}{4}}_{t}$. We show the remaining steps for the left
preconditioner approximation, and the right preconditioner approximation can be
derived similarly.

Since $\hat{L}_{t} \approx {L}^{\frac{-1}{4}}_{t}$, we can say that ${L}_{t}
\approx \hat{L}^{-4}_{t}$, and ${L}_{t-1} \approx \hat{L}^{-4}_{t-1}$. We
substitute ${L}_{t}$ and ${L}_{t-1}$ on both sides of
Equation~\ref{eqn:shampoo-bottleneck}, which gives us:
\begin{align}
     \hat{L}^{-4}_{t} &= \beta_{2} \hat{L}^{-4}_{t-1} + (1-\beta_{2}) G_{t} G^{T}_{t} \\ \nonumber
\implies \hat{L}_{t} &= \left( \beta_{2} \hat{L}^{-4}_{t-1} + (1-\beta_{2}) G_{t} G^{T}_{t} \right)^{\frac{-1}{4}} \\ \nonumber
\end{align}

\begin{align}
\hat{L}_{t} &= \beta^{\frac{-1}{4}}_{2} \hat{L}_{t-1} \left(  I_{m} + \frac{(1-\beta_{2})}{\beta_{2}} \hat{L}^{4}_{t-1} G_{t} G^{T}_{t}  \right)^{\frac{-1}{4}} \\ \nonumber
 &= \beta^{\frac{-1}{4}}_{2} \hat{L}_{t-1} \left(  I_{m} + \frac{(1-\beta_{2})}{\beta_{2}} \eqnmarkbox[Mulberry]{XL}{X_{L}}  \right)^{\frac{-1}{4}} \\ \label{eqn:jorge-rewrite}
\end{align}
\annotate[yshift=-1em]{below,right}{XL}{$\hat{L}^{4}_{t-1} G_{t} G^{T}_{t}$ (line 5, Algorithm 2)}

Next, we get rid of the inverse computation in Equation~\ref{eqn:jorge-rewrite}
by employing the binomial series expansion on the expression in parenthesis.
The binomial theorem for negative exponents suggests that for a square matrix
$A \in \mathbb{R}^{m \times m}$, provided $\| A \| < 1$ and $p > 0$, where $\|
. \|$ is a valid matrix norm, the following is true:
\begin{align}
    \left( I_{m} + A \right)^{-p} &=  \sum_{r=0}^{\infty} ({-1})^{r} \frac{p(p+1)(p+2) ... (p+r-1)}{r!} A^{r} \label{eqn:neg-frac-bin}
\end{align}

Substituting $A = \frac{(1-\beta_{2})}{\beta_{2}} X_{L}$, and $p = \frac{1}{4}$ in Equation~\ref{eqn:neg-frac-bin} yields:
\begin{align}
    \left( I_{m} +  \frac{(1-\beta_{2})}{\beta_{2}} X_{L} \right)^{\frac{-1}{4}} &=  I_{m} - \frac{1}{4} \frac{(1-\beta_{2})}{\beta_{2}} X_{L} + \frac{5}{32} \frac{(1-\beta_{2})^{2}}{\beta^{2}_{2}} X^{2}_{L} + ...  \label{eqn:bin-temp}
\end{align}

Now, replacing the expression in parenthesis in
Equation~\ref{eqn:jorge-rewrite} with its binomial series expansion in
Equation~\ref{eqn:bin-temp}, we remove the inverse calculation entirely as
shown below:
\begin{equation}
    \hat{L}_{t} = \beta^{\frac{-1}{4}}_{2} \hat{L}_{t-1} \left( I_{m} - \frac{1}{4} \frac{ (1-\beta_{2})}{\beta_{2}} X_{L} + \frac{5}{32} \frac{(1-\beta_{2})^{2}}{\beta^{2}_{2}} X^{2}_{L} + ...  \right) \label{eqn:jorge-without-beta}
\end{equation}

Note that the binomial expansion is an infinite series and thus intractable.
In practice, we have found that ignoring the cubic and higher powers of this
expansion does not degrade the sample efficiency of Jorge in comparison to
Shampoo (See Section~\ref{sec:exp-results}).  Hence we drop the higher-order
terms in Equation~\ref{eqn:jorge-without-beta}, which gives us line 6 of
Algorithm~\ref{algorithm:jorge}.  Notice how our preconditioner update step is
composed entirely of matrix-matrix multiplications and additions, which are
highly efficient to compute on GPUs, thereby making Jorge more
compute-efficient than other second-order optimizers.  After updating the
preconditioners, we precondition the gradients by multiplying them with
$\hat{L}_{t}$ and $\hat{R}_{t}$ on the left and right (line 11). Unlike
Shampoo, we do not have to invert our preconditioners because, by definition,
they are an approximation of the inverse fourth roots of Shampoo's
preconditioners.  Finally, the weight update step in lines 14 and 15 is
identical to Shampoo. 

Note that Equation~\ref{eqn:neg-frac-bin} is only valid for $\| A \|<1$, and
therefore for $\| \frac{(1-\beta_{2})}{\beta_{2}} X_{L} \| < 1$.  To ensure
this, Jorge dynamically adjusts $\beta_{2}$ (and $\beta^{\prime}_{2}$ for the
right preconditioner) in each iteration such that the above constraint is met.
We discuss this in detail in Appendix \ref{appendix:binomial-valid}. 

To improve performance, most second-order optimizers, including K-FAC and
Shampoo, typically compute their preconditioners at regular intervals, instead
of every iteration. Following suit, we also allow infrequent preconditioner
updates for Jorge, with the interval kept as a user-configurable
hyperparameter. In the iterations where we do not update the preconditioners,
we simply reuse the preconditioners from the previous iteration. 

As empirical evidence of the efficacy of our approximation we provide the
per-iteration times of SGD, Jorge and AdamW for training
ResNet-50~\citep{resnetCVPR} and DeepLabv3~\citep{deeplabv3} on the Imagenet
and MS-COCO datasets respectively in Table~\ref{tab:compare-batch-times}. For
the ResNet-50 benchmark, we observe that Jorge's iteration times are only 1\%
slower than SGD, whereas it is 26\% faster than Shampoo! For the DeepLabv3 benchmark,
Jorge is only 10\% slower than SGD, but a significant 21\% faster than Shampoo. 

\begin{table}[h]
    \centering 
    \caption{Comparison of wall-clock times per iteration (in seconds) for SGD, Jorge and Shampoo. For Jorge and 
    Shampoo, we compute the preconditioner inverses every 50 iterations, in line with ~\citet{shi2023distributed}.} \label{tab:compare-batch-times}
    {\small
    \begin{tabular}{lrrrrr} \toprule
    Neural Network     & Batch Size  & \# GPUs & SGD     & Jorge & Shampoo \\ \midrule
    ResNet-50          &  1024  & 16 &  0.09 &  0.09 & 0.12 \\ 
    DeepLabv3          &   64   &  4 &  0.33 &  0.37 & 0.47 \\
    \bottomrule
    \end{tabular}
    }
\end{table}

\section{Bootstrapping \jorge's Hyperparameters from SGD}
\label{sec:drop-in}
A new optimizer such as \jorge would be useful in practice only if it does not
require rigorous hyperparameter tuning to achieve a desired level of
generalization on a given training task. Arguably, an important reason behind
the popularity of SGD is the existence of various heuristics for deciding
hyperparameters configurations quickly that can achieve decent generalization.
In this section, we demonstrate \jorge's ability to be an effective drop-in for
SGD. We propose rules to deterministically bootstrap Jorge's hyperparameters
from those of a well-tuned SGD baseline. We call this process ``single-shot
tuning''. There are two implications of being able to single-shot tune Jorge's
hyperparameters from a well-tuned SGD. First, it eliminates the need to explore
the expensive, combinatorial search space of Jorge's hyperparameters.  Second,
the heuristics used to tune SGD's hyperparameters can also be transferred to
\jorge.

Note that we focus on SGD over other adaptive optimizers such as Adam because
prior research has demonstrated that SGD often outperforms adaptive methods in
terms of generalization~\citep{adapative-optimizer-bad-sgd-good,
zhuang2020adabelief, keskar2017improving, luo2019adaptive}. Below, we propose
some rules for transferring SGD's hyperparameters to \jorge.

\textbf{Learning Rate:}
\citet{grafting} propose grafting, a technique for bootstrapping the learning
rate and schedule of a new optimizer from another well-tuned optimizer.
Grafting calculates the magnitude of the weight update by running a step of the
well-tuned optimizer, and the direction of the weight update by running a step
of the new optimizer. Using this approach, we employ grafting to directly use
the learning rate of a well-tuned SGD baseline in \jorge. Integrating grafting
in \jorge involves a small tweak to the weight update step in
Algorithm~\ref{algorithm:jorge} (lines 13-15), which we show in
Appendix~\ref{appendix:jorge-graft}.  However, note that
unlike~\citet{grafting}, we exploit grafting to adopt only the learning rate
from SGD, but not the learning rate schedule (more details below). 

\textbf{Weight Decay Penalty:}
For regularization, in \jorge, we implement the decoupled weight decay scheme
proposed by~\citet{adamw}, as it has been shown to generalize better than L2
regularization for adaptive optimizers. We now explain how the weight decay
penalty for \jorge, $\lambda_{\mathrm{Jorge}}$, can be bootstrapped from SGD.
Let $\beta_{\mathrm{SGD}}$ and $\lambda_{\mathrm{SGD}}$ be the momentum factor
and the weight decay penalty, respectively, of a well-tuned SGD optimizer. We
propose deterministically setting $\lambda_{\mathrm{Jorge}}$ as follows:
\begin{equation}
    \lambda_{\mathrm{Jorge}} = \frac{1}{1-\beta_{\mathrm{SGD}}} \lambda_{\mathrm{SGD}} \label{eqn:jorge-weight-decay}
\end{equation}

Using the almost universal value of 0.9 for $\beta_{\mathrm{SGD}}$, we set
Jorge's weight decay to $10\times$ that of SGD for our experiments.  While
surprisingly simple, we have found this heuristic to work well across several
benchmarks. In Appendix~\ref{appendix:jorge-wd}, we describe the intuition
behind Equation~\ref{eqn:jorge-weight-decay} in more detail.

\begin{figure}[h]
    \centering
    \includegraphics[width=0.49\linewidth]{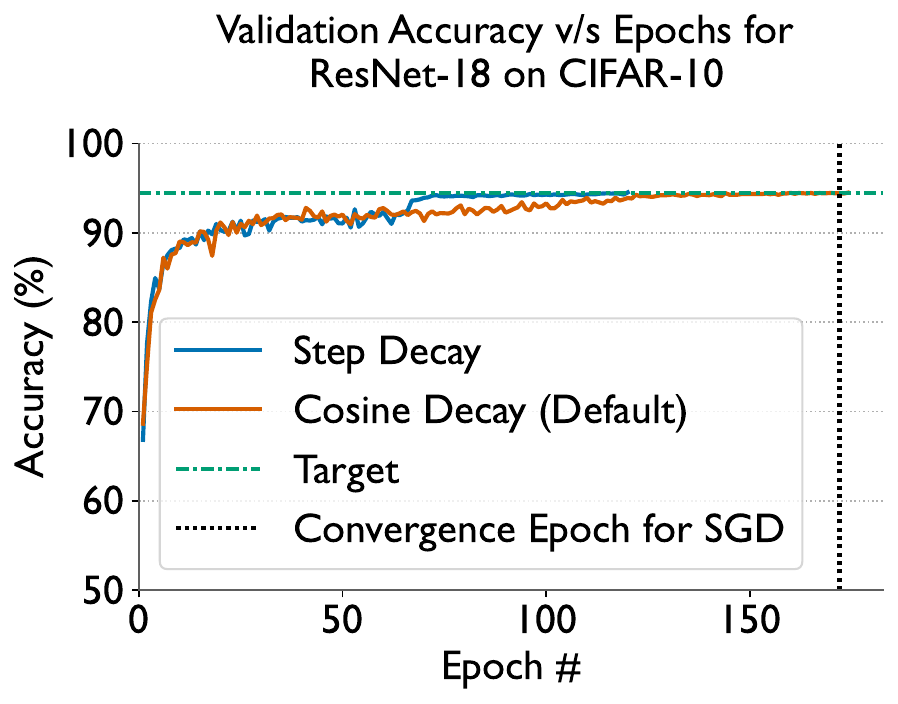}
    \includegraphics[width=0.49\linewidth]{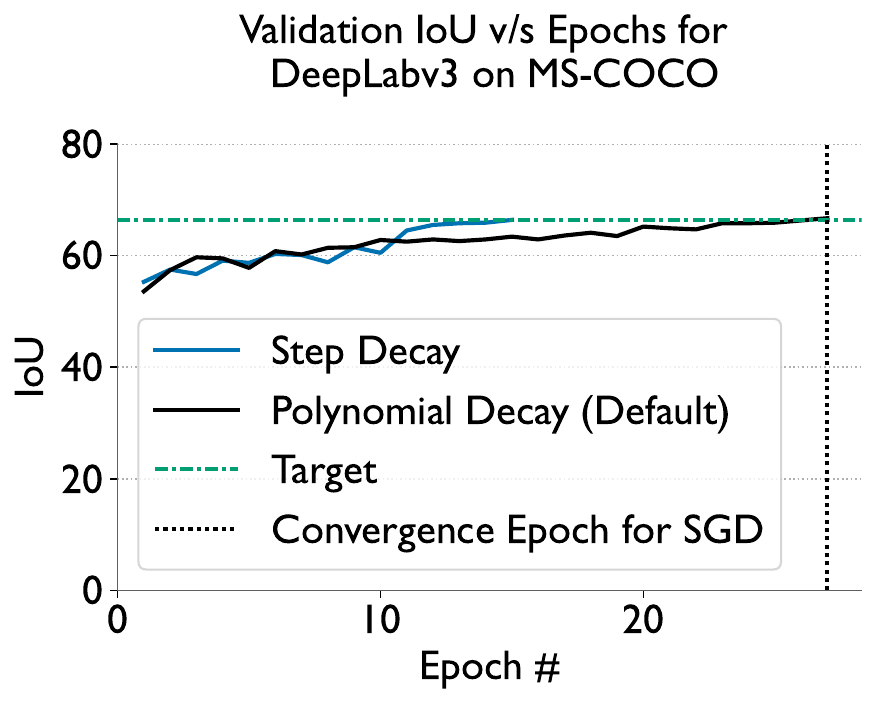}
  \caption{Comparing various learning rate schedules for \jorge. The left and right plots demonstrate the progression of 
  validation accuracy for ResNet-18 on CIFAR-10, and validation IoU for DeepLabv3 on MS-COCO respectively.}
  \label{fig:lr-sched-1}
\end{figure}

\textbf{Learning Rate Schedule}
As per~\citet{grafting}, grafting should allow us to borrow not only the
learning rate, but also the learning rate schedule of a well-tuned SGD
baseline. However, we find that certain learning rate schedules are not
suitable for \jorge. In Figure~\ref{fig:lr-sched-1}, we plot the progression of
validation metrics for training ResNet-18~\citep{he2016identity} on
CIFAR-10~\citep{cifar-10} (left plot) and DeepLabv3~\citep{deeplabv3} on MS
COCO~\citep{mscoco} (right plot).  Note that using the default learning rate
schedules of SGD, which are the cosine~\citep{loshchilov2017sgdr} and
polynomial rate schedules, respectively, leads to barely any improvements in
sample efficiency over SGD. Interestingly, simply switching to the step decay
schedule with 2 decay steps (reducing the learning rate by $10\times$ at each
step) at one-third and two-thirds of the total training epochs (total epochs
same as that of the tuned SGD baseline) resolves this issue. We observe sample
efficiency gains of nearly 1.4--1.8$\times$ over SGD. Therefore, across all
training tasks, we opt for the step decay learning rate schedule with the
aforementioned configuration. Interestingly, in certain scenarios using the
default learning rate schedule of a given well-tuned SGD baseline also leads to
overfitting with Jorge. We discuss this in Appendix~\ref{appendix:jorge-sched}.

\textbf{Preconditioner Update Frequency:}
As mentioned in Section~\ref{sec:formulation}, \jorge has a user-configurable
hyperparameter to  control the frequency at which the preconditioners are
updated.  We suggest using a value for this hyperparameter that brings the
iteration wall-clock times within 10\% of SGD.

\section{Experimental Results}
\label{sec:exp-results}
In this section, we discuss the empirical experiments conducted to evaluate the
efficacy of \jorge against other state-of-the-art optimizers used in deep
learning. 

\subsection{Setup: Benchmarks and Metrics} \label{sec:setup}
Table~\ref{tab:benchmarks} lists the training benchmarks used in our
experiments, all of which are sourced from the torchvision
repository~\citep{torchvision2016}. For each benchmark, we consider two types
of training runs -- one where we let a given optimizer train for the maximum
number of epochs specified in the repository, and the other where we only train
up to the validation metrics specified in Table~\ref{tab:benchmarks}. The former
helps us measure the generalization of each optimizer, whereas the latter helps
us measure the sample efficiencies and total wall-clock times for training.
Mask-RCNN~\citep{mask-rcnn} and DeepLabv3~\citep{deeplabv3} use ResNet-50 as
their backbone. We use SGD as our baseline and also compare with AdamW,
Shampoo, and a recently proposed parallel implementation of
Shampoo~\citep{shi2023distributed}, 

\begin{table}[h]
    \centering 
    \caption{List of benchmarks used to evaluate \jorge against other optimizers. The validation targets for the first two tasks are the 
    same as those used in MLPerf. For the image segmentation task, it is the same as specified in the torchvision repository.} \label{tab:benchmarks}
    {\small
    \begin{tabular}{lllcl}
    \toprule
    Training Task        & Neural Network  & Dataset      & Batch Size(s) & \multicolumn{1}{c}{\begin{tabular}[c]{@{}c@{}}Target Validation\\ Metric \end{tabular}}  \\ \midrule
    Image Classification & ResNet-50        & ImageNet     &  256/1024     &  75.9\% Accuracy  \\ 
    Object Detection     & Mask-RCNN       & MS-COCO 2017 &    32         &  37.7 Bbox mAP \\
    Image Segmentation   & DeepLabv3       & MS-COCO 2017 &    64         &  66.4 IoU \\
    \bottomrule
    \end{tabular}
    }
\end{table}

\textbf{Choice of Hyperparameters:} 
For direct comparisons with SGD and AdamW, we use the default small batch sizes
specified by torchvision, which are 256, 32 and 64 respectively for ResNet-50,
Mask-RCNN, and DeepLabv3. To the best of our knowledge, most evaluations of
second-order optimizers have been conducted at batch sizes much larger than
these values. Thus, to facilitate a direct comparison with Shampoo, we also ran
the ResNet-50 benchmark with a larger batch size of 1024. By doing this, we
could directly borrow the hyperparameters from~\citet{shi2023distributed}, who
evaluated Shampoo in a similar setting.

All the benchmarks from torchvision used in our experiments employ an SGD
optimizer, pre-optimized with a well-calibrated set of hyperparameters.
Accordingly, for our evaluations with SGD, we adhere to these pre-set values.
For our proposed optimizer, \jorge, we adopt the single-shot hyperparameter
configuration outlined in Section~\ref{sec:drop-in}, which is derived directly
from SGD's parameters. We borrow AdamW hyperparameters for the ImageNet
benchmarks from~\citet{heo2021adamp}.  The complete list of all hyperparameters
used in this study can be found in Appendix~\ref{appendix:hparams}.

\textbf{Evaluation Metrics:}
In our evaluation of each benchmark, we record validation accuracy/IoU/mAP with
respect to both number of epochs and wall-clock time. While the epoch-based
measurements provide insights into the sample efficiencies of different
optimizers, wall-clock time offers an understanding of their computational
speed and efficiency on GPU platforms. Together, these metrics offer a
comprehensive assessment of each optimizer's practical efficacy.

\subsection{Comparative Evaluation} \label{sec:results}

Rapid convergence toward a target validation accuracy is not the only goal of
an optimizer. The balance between quick initial convergence and eventual
generalization can dictate an optimizer's selection. For example, SGD remains
the optimizer of choice in computer vision due to its better final validation
accuracy, even though Adam converges faster initially. We evaluate \jorge's
peak validation accuracy against SGD and AdamW across benchmarks, and detail
the results in Table~\ref{tab:smallbs-val}. In these experiments, we let each
optimizer train for the maximum number of epochs specified in the repository.
Notably, for ResNet-50 benchmarks, \jorge exceeds SGD's best validation
accuracy -- 76.02\% vs 76.70\% (large batch size), and 75.97\% -- 76.85\%
(small batch size).  For the Mask-RCNN benchmark, \jorge's IoU of 38.92\%
represents a notable improvement over SGD's 38.3\%.  It's worth highlighting
that these results were achieved using the single-shot tuning strategy
described in Section~\ref{sec:drop-in}.  Though DeepLabv3's performance with
\jorge is marginally worse than that with SGD, the difference is within SGD's
standard deviation, suggesting that small hyperparameter tweaks could bridge
the gap.  Notably, AdamW falls short of SGD's generalization in three out of
four benchmarks but \jorge does better than SGD in three out of four
benchmarks.  Note that this gap in AdamW's generalization compared to SGD has
been a focal point in several prior
studies~\citep{adapative-optimizer-bad-sgd-good, zhuang2020adabelief,
keskar2017improving, luo2019adaptive}.

\begin{table}[h]
    \centering 
    \caption{Maximum validation accuracy ($\mu_{\pm \sigma}$) for SGD, AdamW, and \jorge across benchmarks.} \label{tab:smallbs-val}
    {\small
    \begin{tabular}{lrccccc}
    \toprule
     Neural Network & Batch Size & \# Trials & \# Epochs     &   SGD         & AdamW         & \jorge  \\ \midrule
     ResNet-50     & 1024 & 3 &  90 & 76.02$_{\pm 0.05}$          & 71.85$_{\pm 0.11}$ & \textbf{76.70}$_{\pm 0.07}$    \\
     ResNet-50     & 256 & 3  &  90 & 75.97$_{\pm 0.11}$          & 76.56$_{\pm 0.09}$ & \textbf{76.85}$_{\pm 0.12}$    \\
     DeepLabv3    & 64  & 5  &  30 & \textbf{67.19}$_{\pm 0.16}$ & 66.26$_{\pm 0.20}$ & 67.12$_{\pm 0.12}$    \\
     Mask-RCNN    & 32  & 5  &  26 & 38.30$_{\pm 0.13}$           & 36.58$_{\pm 0.11}$ & \textbf{38.92}$_{\pm 0.10}$   \\
     \bottomrule
    \end{tabular}
    }
\end{table}

\begin{figure}[h]
    \centering
    \includegraphics[width=0.49\linewidth]{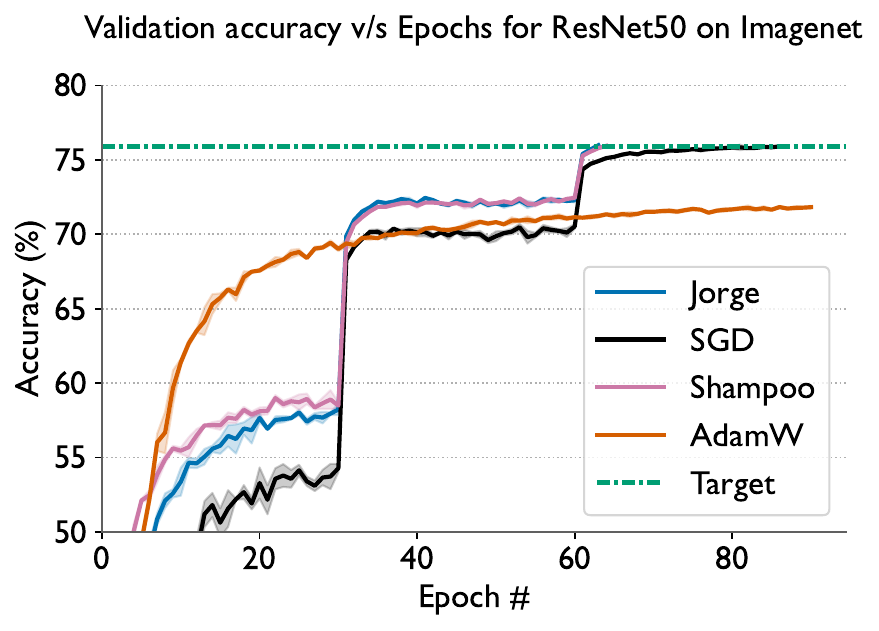}
    \includegraphics[width=0.49\linewidth]{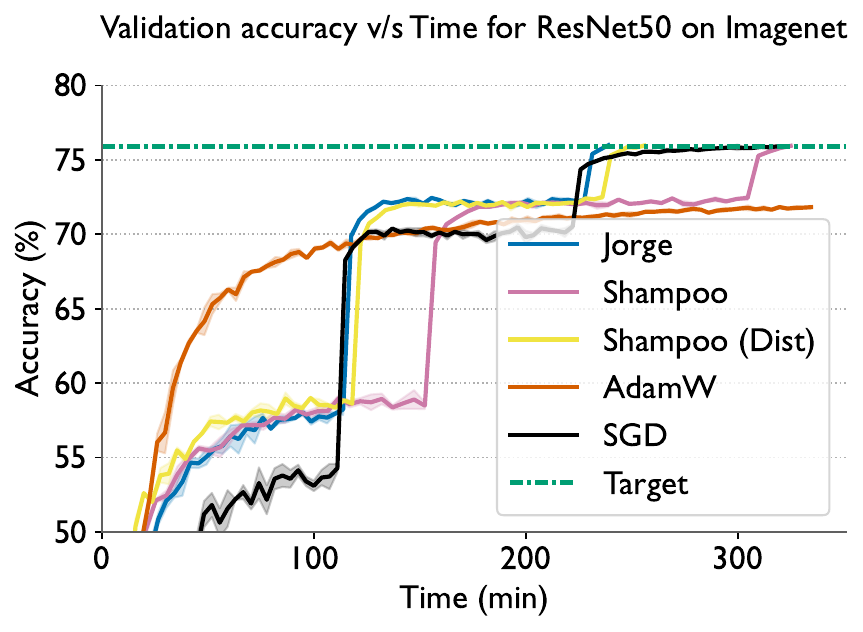}
    \caption{Validation accuracy $\left[\mu \pm \sigma \right]$ v/s epochs (left) and time (right)   
    for the large batch size training (1024) of ResNet-50 on the ImageNet dataset (experiments run on 
    16 A100 GPUs). \label{fig:imagenet-large-bs}} 
\end{figure}

\begin{figure}[h]
    \includegraphics[width=0.33\linewidth]{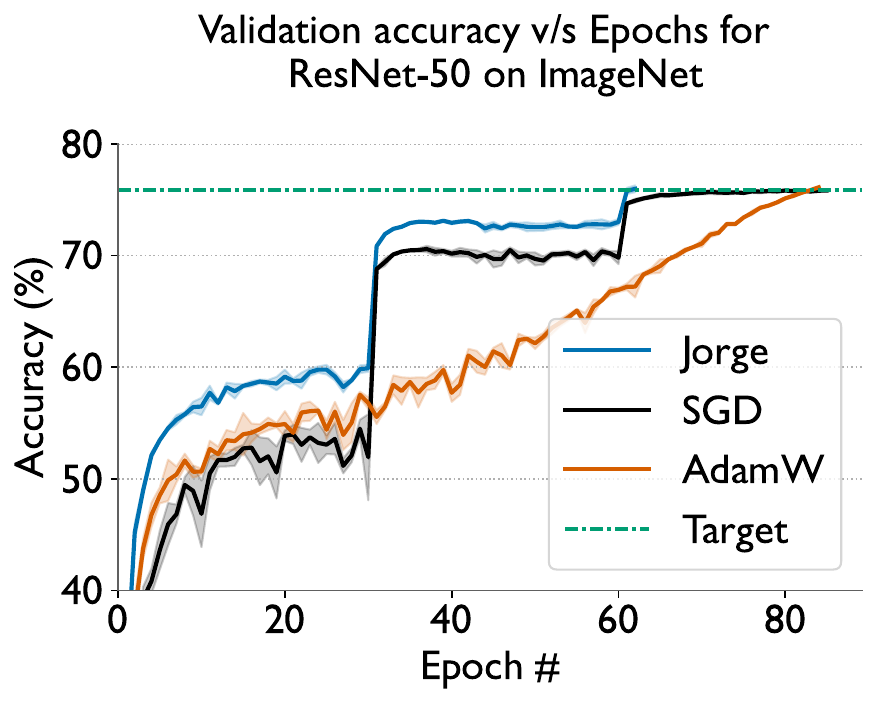}
    \includegraphics[width=0.33\linewidth]{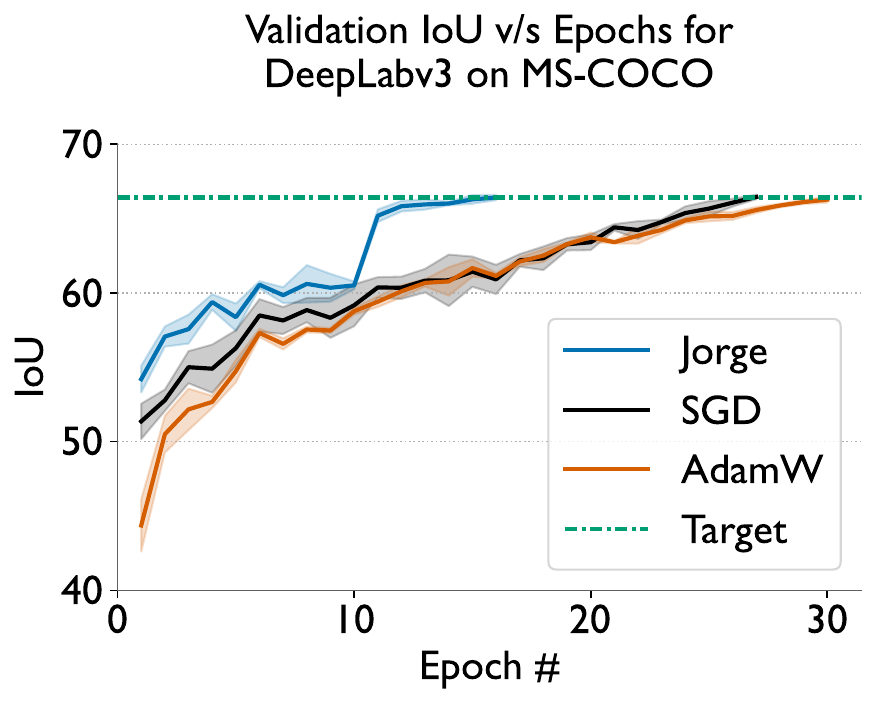}
    \includegraphics[width=0.33\linewidth]{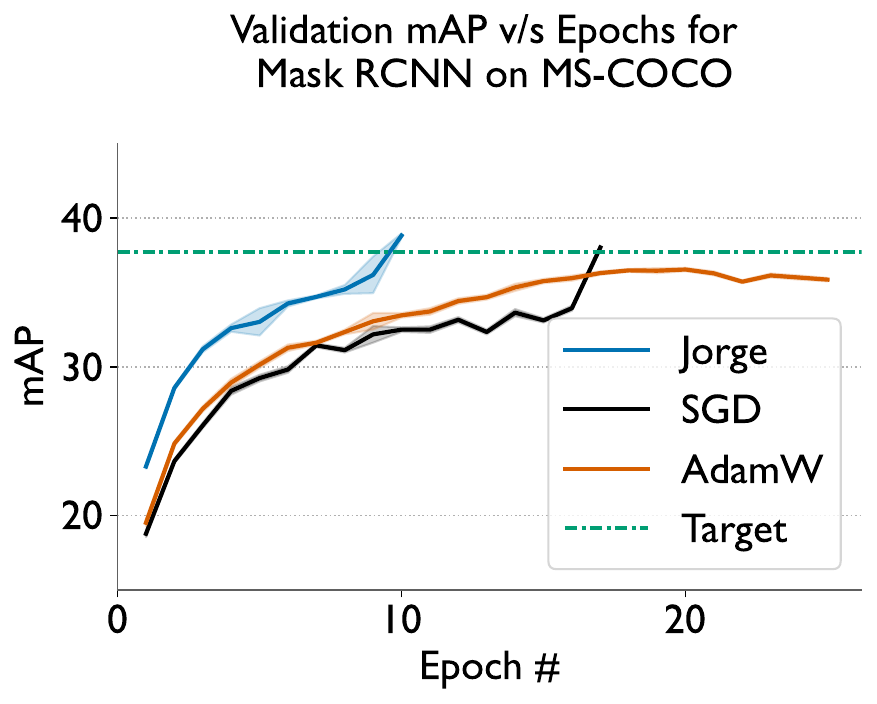}
    \caption{Validation accuracy, IoU, and mAP $\left[\mu \pm \sigma \right]$ v/s epochs for ResNet-50 on ImageNet (left) 
    (batch size of 256), DeepLabv3 on MS-COCO (center), and Mask-RCNN on MS-COCO (right).  \label{fig:smallbs-epochs}}
\end{figure}

Next, we compare the sample efficiency of \jorge to other optimizers.  In this
case, we only train up to the target validation metrics specified in
Table~\ref{tab:benchmarks}. Figure~\ref{fig:imagenet-large-bs} (left) showcases
the progression of validation accuracy over training epochs for ResNet-50 on
ImageNet with the larger batch size of 1024. For other benchmarks, we depict
this progression in Figure~\ref{fig:smallbs-epochs}.  It is evident that in the
context of sample efficiency, \jorge outperforms the first-order optimizers we
compare with -- SGD and AdamW.  Across both the small (256) and large (1024)
batch size training scenarios for ResNet-50, \jorge outperforms SGD by
requiring around 27\% fewer iterations to reach the target validation accuracy
of 75.9\%. The improvements in sample efficiency over SGD across other
benchmarks are markedly higher -- 40\% for DeepLabv3, and 41\% for Mask-RCNN.
Again, we achieve these results by simply bootstrapping \jorge's
hyperparameters from SGD, only making the changes outlined in
Section~\ref{sec:drop-in}.  The improvements in sample efficiency over AdamW
are similar to those over SGD. Also, AdamW falls short of achieving the target
validation metric in two out of four experiments.

As discussed in Section~\ref{sec:formulation}, we have designed \jorge to
approximate Shampoo with a focus on GPU efficiency.
Figure~\ref{fig:imagenet-large-bs} (left) demonstrates that \jorge achieves the
target validation accuracy in almost the same number of epochs as Shampoo (62
vs.~63). This observation strongly validates our approach and confirms that
\jorge's approximations do not degrade its statistical efficiency.

Let us now turn our attention to an equally crucial metric: wall-clock time
required for training. Figure~\ref{fig:imagenet-large-bs} (right) demonstrates
the progression of validation accuracy over time for the large batch size
training of ResNet-50. We observe that \jorge achieves the target validation
accuracy in 25\% less time compared to SGD, which is a significant improvement.
If we consider the serial implementation of Shampoo (pink line), it takes more
total time to converge than SGD despite requiring 27\% fewer epochs.  This
observation demonstrates the prowess of \jorge as a GPU-efficient adaptation of
Shampoo: it's significantly faster than Shampoo's wall-clock time for
convergence (239 minutes vs.~325 minutes), despite requiring a similar number
of epochs. As noted in Section~\ref{sec:related}, the prevailing approach for
mitigating the large overhead of preconditioning has been to develop
distributed implementations of these optimizers. Within this context,
Figure~\ref{fig:imagenet-large-bs} (right) also presents the wall-clock time of
a state-of-the-art parallel implementation of Shampoo (yellow
line)~\citep{shi2023distributed}.  Notably, even though \jorge executes locally
on each GPU, it still manages to yield a 4\% speedup over the parallel version
of Shampoo.

While a 4\% improvement might seem modest, its implications are more
far-reaching. Often times, AI practitioners do not have access to large numbers
of GPU resources. In such resource-constrained settings, \jorge might be an
ideal optimizer when parallelizing across GPUs is not an option. This also
applies to environments with limited interconnect bandwidth.  

Finally, we focus on the small batch size benchmarks to evaluate how \jorge's
training wall-clock times compare with other first-order optimizers. We present
these results in Table~\ref{tab:smallbs-time}.  Once again, \jorge makes
significant improvements in the total training wall-clock times. Compared to
SGD, \jorge improves the time to convergence by 23\%, 34\%, and 45\% for
ResNet-50, DeepLabv3, and Mask-RCNN respectively. The corresponding
improvements over AdamW are even higher -- 26\%, 41\%, and 58\% (the last
number is much higher since AdamW did not converge on that run). The wall-clock
time improvements in these experiments highlight \jorge's applicability to
small batch size training scenarios, where the overheads of a second-order
optimizer cannot be masked behind network computation, making it more
challenging for \jorge to beat first-order optimizers.

\begin{table}[h]
    \centering 
    \caption{ Comparison of the total training time (in minutes) of \jorge with SGD and AdamW for the small batch size benchmarks (experiments run on 
    four A100 GPUs).} \label{tab:smallbs-time}
    {\small
    \begin{tabular}{lrcrrr}
    \toprule
     Neural Network & Batch Size & \# Trials     &   SGD         & AdamW         & \jorge  \\ \midrule
     ResNet-50       & 256 & 3 & 1005$_{\pm 40}$ & 1052$_{\pm 36}$  &  \textbf{781}$_{\pm 44}$ \\
     DeepLabv3      & 64  & 5 & 217$_{\pm 12}$  & 244$_{\pm 01}$ &  \textbf{144}$_{\pm 30}$ \\
     Mask-RCNN      & 32  & 5 & 332$_{\pm 47}$  & 438$_{\pm 14} $  &  \textbf{182}$_{\pm 11}$ \\
     \bottomrule
    \end{tabular}
    }
\end{table}

\section{Conclusion and Future Work}
\label{sec:conc}
In this work, we introduced \jorge, an efficient, adaptive, second-order
optimizer tailored to GPU platforms. We eliminated the primary computational
bottleneck of computing matrix inverses in second-order optimizers by proposing
a novel approximation of the preconditioner computation in Shampoo, which
sidesteps the need to explicitly compute matrix inverses. Further, we proposed
a single-shot hyperparameter tuning strategy, that can directly bootstrap
\jorge's hyperparameters from a well-tuned SGD baseline without the need to
conduct extensive tuning. We evaluated \jorge against state-of-the-art
first-order optimizers -- SGD and AdamW, as well as Shampoo, and we
demonstrated improvements in generalization, sample efficiencies, and training
wall-clock times. As future work, we plan to develop a single-shot
hyperparameter bootstrapping strategy from AdamW as well. This will allow us to
employ \jorge to train large language models. Additionally, we plan to develop
a distributed implementation of \jorge to reduce its per-GPU memory
consumption, which currently stands at 1.5--2$\times$ that of Adam (see
Appendix~\ref{appendix:jorge-memory-consumption}).


\vspace{0.1in}
\noindent{\bf Reproducibility Statement:}
We are committed to enabling reproducibility of our work, as it ensures correct
and transparent results.  We plan to open source the code for \jorge as well as
the benchmarks evaluated in this paper.  Additionally, we provide a
comprehensive list of all hyperparameters used in this study for each optimizer
and each benchmark in Appendix~\ref{appendix:hparams}. The hyperparameters can
be directly substituted as  the arguments of SGD and AdamW shipped with PyTorch
2.0 in the ``torch.optim'' package. Similarly, the hyperparameters listed for
Jorge will be compatible with our open source codebase.

\bibliographystyle{mlconf}
\bibliography{./bib/cite,./bib/pssg}

\newpage
\appendix
\section{Appendix}
\subsection{Ensuring validity of the binomial expansion by dynamically
adjusting $\beta_{2}$}
\label{appendix:binomial-valid}

In Section~\ref{sec:formulation}, we mentioned that for the binomial expansion
in Equation~\ref{eqn:jorge-without-beta} to be valid, we must also ensure that
$\left \| \frac{(1-\beta_{2})}{\beta_{2}} X_{L} \right \| < 1$. To ensure this
condition is met at every iteration, Jorge dynamically updates the EMA update
parameters $\beta_{2}$ and $\beta^{\prime}_{2}$ (for the right preconditioner)
at each iteration. We start with the condition we want to ensure and derive a
lower bound on $\beta_{2}$.
\begin{align}
    \left \| \frac{(1-\beta_{2})}{\beta_{2}} X_{L} \right \| < 1 \implies \beta_{2} > \frac{\| X_{L} \|}{\| X_{L} \| + 1}  \label{eqn:beta}
\end{align}

Therefore, we need to set $\beta_{2}$ to a value higher than $\frac{\| X_{L}
\|}{\| X_{L} \| + 1}$  to ensure the validity of the binomial expansion.  In
practice, we have seen that setting $\beta_{2}$ equal to this quantity works
well, provided we are using the Frobenius norm as our matrix norm function of
choice.

Substituting the value of $\beta_{2}$ from Equation~\ref{eqn:beta} in
Equation~\ref{eqn:jorge-without-beta} and ignoring the cubic and higher powers,
gives us the complete left preconditioner update step:
\begin{equation}
    \hat{L}_{t} = \left( \frac{ \| X_{L}\| + 1}{\| X_{L} \|} \right)^{\frac{1}{4}} \hat{L}_{t-1} \left( I_{m} - \frac{1}{4} \frac{X_{L}}{\| X_{L} \|} + \frac{5}{32} \frac{X^{2}_{L}}{\| X_{L} \|^{2}} \right) \label{eqn:jorge-final-update-rule}
\end{equation}

The corresponding formulation of  $\beta^{\prime}_{2}$ for the right
preconditioners can be derived in a similar manner.

\subsection{Jorge with grafting}
\label{appendix:jorge-graft}

In Section~\ref{sec:drop-in}, we mentioned adding grafting to Jorge, which adds
a step in the weight update step of Algorithm~\ref{algorithm:jorge}.  Grafting
maintains the direction of the current step ($\frac{m_{t}} {\| m_{t} \|}$), but
uses the magnitude of the step of a well-tuned optimizer
($\|m_{\mathrm{SGD},t}\|$ in this case).  In
Algorithm~\ref{algorithm:jorge-sgd-graft} below, we see $m_{t}$ becomes
$\|m_{\mathrm{SGD},t}\| \frac{m_{t}}{\| m_{t}\|}$.
\begin{algorithm}[H]
    \centering
    \caption{Jorge's modified weight update rule with SGD grafting}\label{algorithm:jorge-sgd-graft}
    \begin{algorithmic}[1]
        \State \textbf{Update Weights:}
        \State $m_{t} = \beta_{1} m_{t-1} + (1-\beta_{1}) \hat{G_{t}}$          \Comment{Jorge weight update}
        \State $m_{\mathrm{SGD},t} = \beta_{1} m_{\mathrm{SGD},t-1} + G_{t}$     \Comment{SGD (with heavy ball momentum) weight update}
        \State
        \State $\theta_{t} = \theta_{t} - \eta_{t}  \| m_{\mathrm{SGD},t} \| \frac{m_{t}}{\| m_{t} \|}$ \Comment{Grafted weight update}
    \end{algorithmic}
\end{algorithm}

\begin{figure}[t]
    \includegraphics[width=0.49\textwidth]{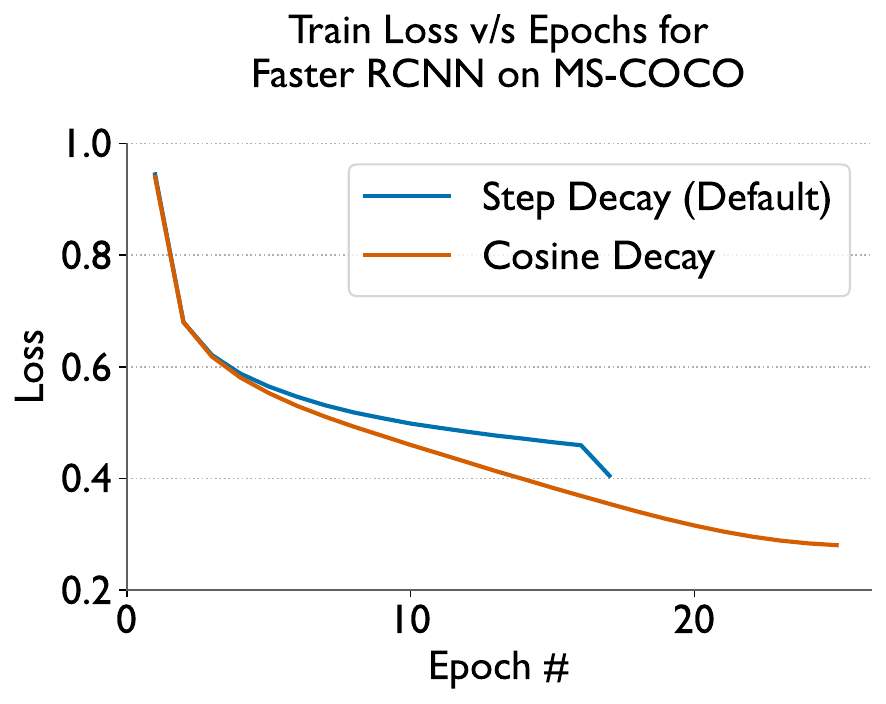}
    \includegraphics[width=0.49\textwidth]{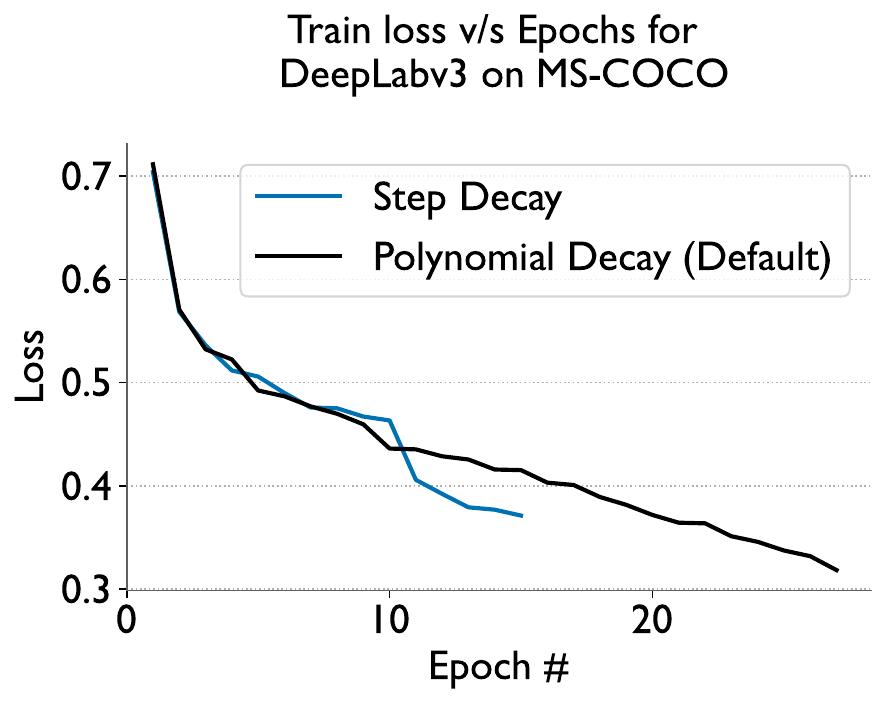}
    \includegraphics[width=0.49\textwidth]{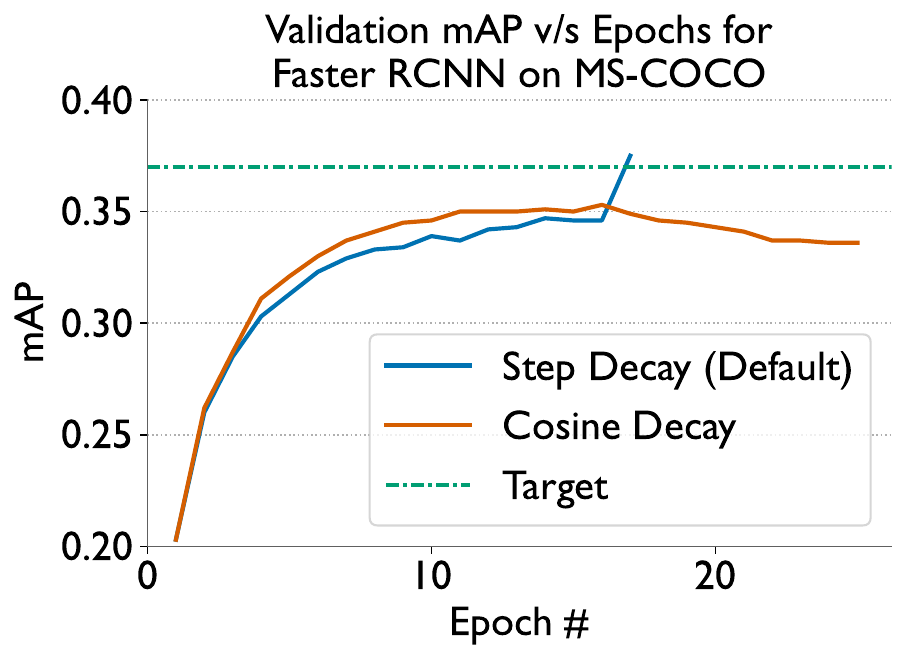}
    \includegraphics[width=0.49\textwidth]{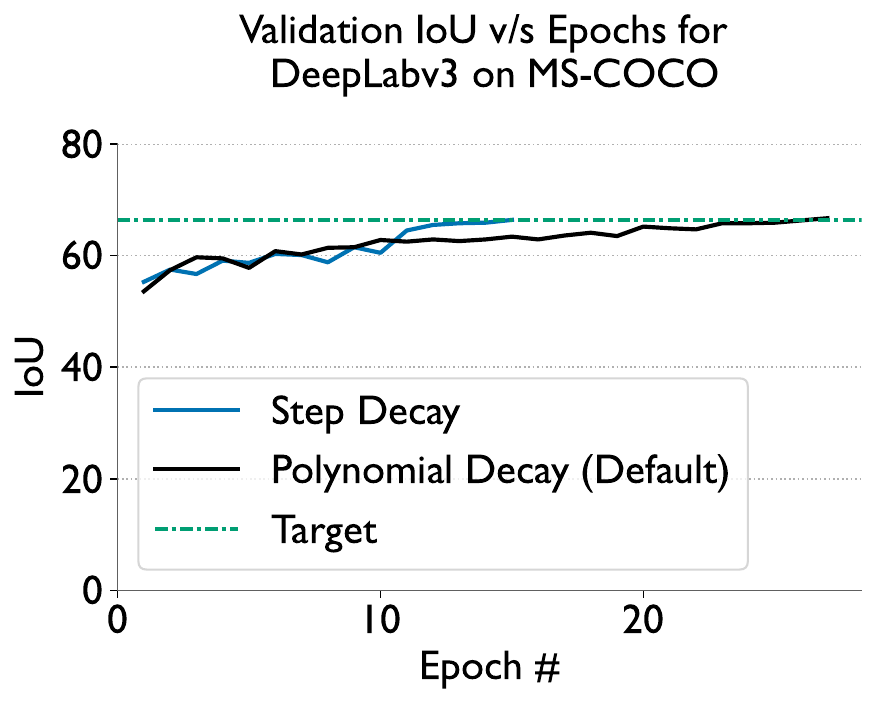}
  \caption{Comparison of different learning rate schedules with Jorge on two training tasks - Object Detection 
  with Faster RCNN~\citep{faster_rcnn}, and Image Segmentation with DeepLabv3~\citep{deeplabv3}, both on the 
  MS-COCO dataset~\citep{mscoco}. Both training tasks use a batch size of 64.}
\label{appendix-fig:lr}
\end{figure}

\subsection{Intuition behind Jorge's weight decay heuristic}
\label{appendix:jorge-wd}

As mentioned in Section~\ref{sec:drop-in}, we created a simple heuristic for
setting Jorge's weight decay.  Let the parameters at time step $t$ be
$\theta_{t}$. SGD will first calculate the weight decay update as
$\lambda_{\mathrm{SGD}}\theta_{t}$ and then update its running estimate of the
momentum using the gradients from the loss and
$\lambda_{\mathrm{SGD}}\theta_{t}$.
Since the weight decay is a part of the running momentum estimates, the weight
decay calculated at time step $t$ will influence the parameter updates at time
step $t+\tau$, albeit attenuated by $\beta_{\mathrm{SGD}}^{\tau}$. Therefore,
the \emph{effective} contribution of a weight decay update calculated at time
step $t$ is:
\begin{equation}
    \sum_{\tau=0}^{T-t} \beta_{\mathrm{SGD}}^{\tau} \lambda_{\mathrm{SGD}} \theta_{t} \approx \frac{1}{1-\beta_{\mathrm{SGD}}} \lambda_{\mathrm{SGD}} \theta_{t}
\end{equation}

Since we use a decoupled weight decay scheme for Jorge, the weight decay
calculated at time step $t$ does not contribute to future weight updates.
Therefore, to match the effective contribution of the weight decay updates in
SGD, we set the weight decay penalty for Jorge to
$\frac{1}{1-\beta_{\mathrm{SGD}}} \times$ that of SGD, as shown in
Equation~\ref{eqn:jorge-weight-decay}.

\subsection{Experiments with learning rate schedules}
\label{appendix:jorge-sched}

Here we discuss the phenomenon of certain learning rate schedules leading to
overfitting with Jorge, that we briefly alluded to in
Section~\ref{sec:drop-in}. Figure~\ref{appendix-fig:lr} (left) demonstrates the
training loss and validation mAP curves for Faster-RCNN on MS-COCO. Notice that
while the cosine schedule never reaches the target validation mAP, this is not
because it sufficiently fails to minimize the training loss. Infact, it leads
to a training loss significantly lower than the step decay schedule, thereby
indicating overfitting.

Similarly, Figure~\ref{appendix-fig:lr} (right) demonstrates the training loss
and validation IoUs for the image segmentation task with DeepLabv3. Here, the
polynomial-scheduled Jorge must reach a lower training loss (loss of $0.32$)
than the stepwise-scheduled Jorge ($0.37$) to reach the same validation
accuracy, once again symbolizing overfitting.

Our hypothesis for the phenomenon is that Jorge requires a high learning rate
in the initial phases of training to escape sharp local minima. Due to its more
accurate updates it is more prone towards falling into sharp minima compared to
SGD, which might escape these because of its noisy updates. We plan to explore
this phenomenon in more detail in future work. 

\subsection{List of Hyperparameters for Section~\ref{sec:exp-results}}
\label{appendix:hparams}

We list the hyperparameters used in this study for SGD, Jorge, and AdamW in
Tables~\ref{sgd-hp}, ~\ref{jorge-hp}, and~\ref{adamw-hp} respectively. For
Shampoo, we have used the same learning rate, weight decay and learning rate
schedule as SGD, as per the recommendation of~\citet{shi2023distributed} and
enabled SGD grafting. 

\begin{table}[h]
    \centering 
    \caption{Hyperparameters used in this study for SGD. These are the defaults in torchvision. \label{sgd-hp}}
    {\small
    \begin{tabular}{lllll} \toprule
    Hyperparameter        & \begin{tabular}[c]{@{}l@{}}Resnet-50\\ (batch size 1024)\end{tabular}                                          & \begin{tabular}[c]{@{}l@{}}ResNet-50\\ (batch size 256)\end{tabular}   & DeepLab-v3                                                                 & Mask RCNN                                                                    \\ \midrule
    Learning Rate         & 0.4                                                                                                              & 0.1                                                                      & 0.02                                                                       & 0.02                                                                         \\
    Weight Decay          & $1\mathrm{e}{-4}$                                                                                                & ${1\mathrm{e}{-4}}$                                                      & $1\mathrm{e}{-4}$                                                          & $1\mathrm{e}{-4}$                                                            \\
    \begin{tabular}[c]{@{}l@{}}Learning Rate\\ Schedule\end{tabular} & \begin{tabular}[c]{@{}l@{}}Linear warmup over\\5 epochs. Then step\\decay at epochs 30\\and 60\end{tabular} & \begin{tabular}[c]{@{}l@{}}Step decay at\\epochs 30 and 60\end{tabular} & \begin{tabular}[c]{@{}l@{}}Polynomial decay\\with 0.9 power\end{tabular} & \begin{tabular}[c]{@{}l@{}}Step decay at\\epochs 16 and 22\end{tabular} \\ 
    Momentum & 0.9 & 0.9 & 0.9 & 0.9 \\ 
    Nesterov & False & False & False & False \\ \bottomrule
    \end{tabular}
    }
\end{table}

\begin{table}[h]
    \centering 
    \caption{Hyperparameters used in this study for Jorge. \label{jorge-hp}} 
    {\small
    \begin{tabular}{lllll} \toprule
    Hyperparameter        & \begin{tabular}[c]{@{}l@{}}Resnet-50\\ (batch size 1024)\end{tabular}                                          & \begin{tabular}[c]{@{}l@{}}ResNet-50\\ (batch size 256)\end{tabular}   & DeepLab-v3                                                                 & Mask RCNN                                                                    \\ \midrule
    Learning Rate         & 0.4                                                                                                              & 0.1                                                                      & 0.02                                                                       & 0.02                                                                         \\
    Weight Decay          & $1\mathrm{e}{-3}$                                                                                                & ${1\mathrm{e}{-3}}$                                                      & $1\mathrm{e}{-3}$                                                          & $1\mathrm{e}{-3}$                                                            \\
    \begin{tabular}[c]{@{}l@{}}Learning Rate\\ Schedule\end{tabular} & \begin{tabular}[c]{@{}l@{}}Linear warmup over\\5 epochs. Then step\\ decay at epochs 30\\and 60\end{tabular} & \begin{tabular}[c]{@{}l@{}}Step decay at\\epochs 30 and 60\end{tabular} & \begin{tabular}[c]{@{}l@{}}Step decay at\\epochs 10 and 20\end{tabular} & \begin{tabular}[c]{@{}l@{}}Step decay at\\epochs 8 and 16\end{tabular} \\ 
    Momentum & 0.9 & 0.9 & 0.9 & 0.9 \\ 
    \begin{tabular}[c]{@{}l@{}}Preconditioner \\ Update Freq.\end{tabular} & 50 & 2 & 4 & 8 \\ \bottomrule
    \end{tabular}
    }
\end{table}

\subsection{Analysis of memory consumption}
\label{appendix:jorge-memory-consumption}

We mentioned in Section~\ref{sec:conc} that Jorge consumes $1.5-2\times$ the
memory of Adam.  This is because Adam uses 2 32-bit floating point optimizer
states per parameter.  In contrast, Jorge uses 3 (hence $1.5\times$), one each
for the left preconditioner, right preconditioner, and momentum (see
Algorithm~\ref{algorithm:jorge}). It becomes 4 once grafting is introduced
(hence $2\times$), due to the fact that we now need to maintain the momentum
for SGD as well. This is a major limitation of our method, and one which we
plan to fix with a distributed implementation.

\begin{table}[h]
    \centering
    \caption{Hyperparameters used in this study for AdamW. \label{adamw-hp}}
    {\small
    \begin{tabular}{lllll} \toprule
    Hyperparameter        & \begin{tabular}[c]{@{}l@{}}Resnet-50\\ (batch size 1024)\end{tabular}                                          & \begin{tabular}[c]{@{}l@{}}ResNet-50\\ (batch size 256)\end{tabular}   & DeepLab-v3                                                                 & Mask RCNN                                                                    \\ \midrule
    Learning Rate         & 0.004                                                                                                              & 0.001                                                                      & 0.0002                                                                     & 0.0002                                                                         \\
    Weight Decay          & 0.1                                                                                                & 0.1                                                     & $1\mathrm{e}{-2}$                                                          & $1\mathrm{e}{-2}$                                                            \\
    \begin{tabular}[c]{@{}l@{}}Learning Rate \\ Schedule\end{tabular} & Cosine & Cosine & Cosine & Cosine\\ 
    Momentum & 0.9 & 0.9 & 0.9 & 0.9 \\ 
    $\beta$s & (0.9, 0.999) & (0.9, 0.999) & (0.9, 0.999) & (0.9, 0.999) \\ 
    $\epsilon$ & $1\mathrm{e}{-8}$ & $1\mathrm{e}{-8}$ & $1\mathrm{e}{-8}$ & $1\mathrm{e}{-8}$ \\ 
    amsgrad   & False & False & False & False \\ \bottomrule
    \end{tabular}
    }
\end{table}

\end{document}